# Structured Gradient Guidance for Few-Shot Adaptation in Large Language Models

Hongye Zheng
The Chinese University of Hong Kong
Hong Kong, China

Yichen Wang
Georgia Institute of Technology
Atlanta, USA

Ray Pan
Independent Researcher
Seattle, USA

Guiran Liu
San Francisco State University
San Francisco, USA

Binrong Zhu
San Francisco State University
San Francisco, USA

Hanlu Zhang*
Stevens Institute of Technology
Hoboken, USA

*Abstract*-This paper presents a gradient-informed fine-tuning method for large language models under few-shot conditions. The goal is to enhance task adaptability and training stability when data is limited. The method builds on a base loss function and introduces two gradient-related regularization terms. The first enforces gradient direction consistency to guide parameter updates along task-relevant directions and prevent drift. The second controls gradient magnitude to avoid abnormal updates. Together, these components support a more efficient and stable optimization path. To further improve cross-task generalization, the method incorporates a gradient alignment mechanism. This mechanism measures the consistency between optimization directions of the source and target tasks. It enhances the model's fine-tuning performance in multi-task and cross-domain scenarios. Across various natural language understanding tasks, the method outperforms existing fine-tuning strategies in average accuracy, gradient stability, and directional alignment. Empirical evaluations under different sample sizes and domain-specific tasks confirm the method's robustness and broad applicability in low-resource environments. In particular, the method shows clear advantages in controlling parameter update paths. The results demonstrate that a gradient-based fine-tuning framework can effectively leverage the representational power of large language models. It ensures training stability while reducing dependence on large volumes of labeled data.

I. INTRODUCTION

In recent years, Large Language Models (LLMs) have achieved remarkable success in natural language processing tasks, driving intelligent upgrades in many downstream applications. With strong representational and contextual understanding capabilities, LLMs have become the backbone of intelligent system development[1]. However, as model sizes continue to grow, efficiently adapting them to new tasks and domains has become an urgent challenge. This issue is particularly prominent in low-resource scenarios, where traditional full-parameter fine-tuning methods are costly and prone to overfitting and poor generalization [2-3]. As a result, developing efficient fine-tuning methods under limited data conditions has become a critical research topic in artificial intelligence [4].

Fine-tuning strategies play a pivotal role in addressing few-shot learning problems. Existing methods often rely on structural modifications to pretrained models or parameter-efficient tuning of submodules [5]. However, they tend to overlook the value of gradient information in guiding optimization. Gradients indicate sensitive directions in parameter space and reveal how the model adjusts toward task objectives. These properties connect the model's learning capacity with task adaptability. In few-shot learning, the sparsity, directionality, and locality of gradients become more pronounced. Effectively leveraging these features can improve learning efficiency while maintaining model stability.

Moreover, introducing gradient information helps mitigate the lack of generalization in low-data settings. With limited training samples, models struggle to capture deeper semantic relationships. Gradient-based adjustments can guide the model toward more suitable paths in parameter space, enhancing task-specific representations. This optimization-driven strategy not only improves task alignment but also reduces the reliance on large annotated datasets. This is especially valuable in practical domains such as healthcare, law, and finance, where labeled data is both scarce and expensive [6-8]. As such, fine-tuning techniques for few-shot learning face higher demands for practical effectiveness[9].

Given the increasing structural complexity of LLMs, exploring gradient-based optimization strategies is crucial for building more general and robust fine-tuning frameworks. Compared to static tuning methods, dynamically sensing and utilizing gradient signals allows finer control and better guidance during parameter updates. This can accelerate adaptation to specific tasks and improve the stability and consistency of fine-tuning outcomes. Gradient-driven tuning also offers new perspectives for research in model safety and fairness. By adjusting sensitive gradient directions, it may be possible to suppress bias propagation and abnormal activations [10]. In summary, optimizing few-shot fine-tuning strategies for LLMs through gradient information holds both theoretical and practical value. This direction aligns with the broader pursuit of efficiency and sustainability in AI systems. It also facilitates a key transition from general capabilities to task specialization in large models [11-12]. As model sizes grow and task complexity increases, releasing model potential under resource constraints will be a key factor in next-generation intelligent system design. Therefore, gradient-informed fine-tuning represents a critical step toward addressing future AI challenges.

## II. METHOD

This study proposes a small-sample fine-tuning method for large language models that incorporates gradient information to optimize the parameter update trajectory. The central idea is to embed gradient-based characteristics—specifically direction and magnitude—into the fine-tuning process to enhance the model's adaptability under extreme data scarcity. While conventional fine-tuning approaches often rely on direct loss minimization, they tend to overlook the geometric and structural insights contained within gradient dynamics. Cai et al. [13] demonstrated the effectiveness of dynamically regulating low-rank adaptation paths in few-shot scenarios, highlighting the importance of controlled parameter updates in maintaining model robustness. Similarly, Zhu et al. [14] emphasized structured preference modeling in reinforcement learning-based fine-tuning, underlining the value of incorporating task-relevant guidance in model optimization. Drawing further inspiration from Wang [15], who employed semantic alignment via LLama-based modeling for controlled generalization, our method integrates gradient-oriented regularization to guide parameter changes along semantically consistent directions. This approach not only stabilizes the optimization process but also facilitates better generalization across tasks. The overall model architecture is depicted in Figure 1.

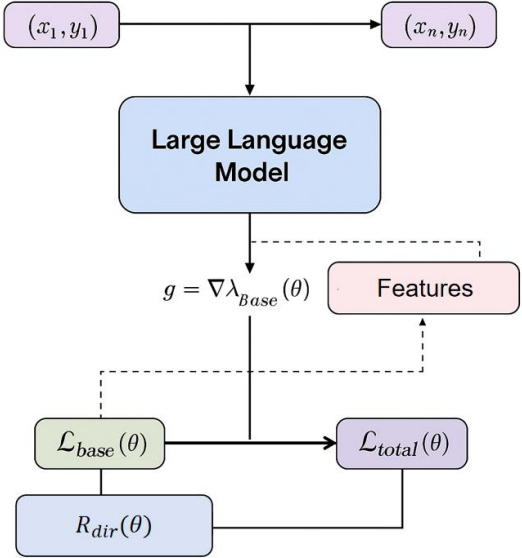

Figure 1. Overall model architecture diagram

This model architecture diagram shows how to optimize the fine-tuning process of a large language model based on gradient information under the condition of few samples. First, the gradient $g = \nabla_\theta L_{base}(\theta)$ is calculated through the basic loss function, and then it is used to construct the direction regularization term $R_{dir}(\theta)$ and other feature control terms to guide the direction and amplitude of parameter adjustment. Finally, the losses of each part are aggregated into the total loss $L_{total}(\theta)$ to achieve gradient-aware optimization of the model fine-tuning path.

This method controls the sensitivity of the model in a specific dimension and enhances the focus on important features by explicitly modeling the gradient direction and magnitude. In the initialization stage, we define the basic loss function as:

$$L_{base}(\theta) = \frac{1}{N}\sum_{i=1}^{N} l(f_\theta(x_i), y_i)$$

Where $\theta$ represents the model parameters, $f_\theta$ is the large language model, $l$ is the task-specific loss function (such as cross entropy), and $(x_i, y_i)$ is the input-output pair. In order to guide the optimization process along the more informative gradient direction, we calculate the gradient of the basic loss function with respect to the parameters:

$$g = \nabla_\theta L_{base}(\theta)$$

To further refine the optimization trajectory, a gradient direction regularization term is introduced. This term functions to suppress updates in directions dominated by noise while amplifying those aligned with the principal task-relevant gradient. By emphasizing directional coherence, it ensures that parameter changes contribute meaningfully to model adaptation rather than diverging due to stochastic fluctuations. Wang [16] demonstrated how topology-aware signal filtering in multi-agent systems improves decision-making consistency by suppressing noisy influence zones, a concept that aligns closely with our intent to stabilize update directionality. Cheng [17] further highlighted the importance of structured directional learning through automated feature extraction, which informs our design of directional regularization to capture dominant trends in limited data. Moreover, Gong et al. [18] showed that semantic coherence in fine-tuning pipelines—particularly in high-risk domains like financial fraud—relies on reinforcing consistent optimization paths. Motivated by these insights, we formulate the gradient direction regularization term as:

$$R_{dir}(\theta) = \lambda_1 \| \frac{g}{\|g\|} - d_{prior} \|^2$$

Where $d_{prior}$ is the reference gradient direction (which can be the average gradient direction in the pre-training phase or the prior task direction), and $\lambda_1$ is a hyperparameter for adjusting the intensity. This regularization term can encourage the model to learn in a consistent direction, thereby enhancing cross-task generalization capabilities.

In addition, under low resource conditions, the gradient amplitude has a direct impact on the model update. For this reason, we introduce a gradient amplitude control term to control the update amplitude by weight to alleviate the risk of overfitting.

$$R_{dir}(\theta) = \lambda_2 (\| g \| - \tau)^2$$

$\tau$ is the set gradient amplitude target, and $\lambda_2$ is a hyperparameter that adjusts the penalty strength of this term. This regularization term can enhance the robustness of the model in the early training stage and prevent it from over-adjusting under noisy gradients. In the total loss function, we integrate the above two regularization terms into the basic loss to obtain the final optimization target:

$$L_{total}(\theta) = L_{base}(\theta) + R_{dir}(\theta) + R_{mag}(\theta)$$

In order to further improve the adaptability of few samples, this method also introduces a gradient contrast mechanism to measure the consistency of gradient distribution between different tasks to dynamically adjust the fine-tuning strength. Specifically, let $g^{(t)}$ and $g^{(s)}$ be the gradients of the current target task and the source task respectively, and we define the gradient contrast loss as:

$$R_{grad} = \lambda_3 \cdot (1 - \cos < g^{(t)}, g^{(s)} >)$$

Where $\cos < \cdot, \cdot >$ represents the cosine similarity of the two vectors, and $\lambda_3$ is the weight factor. This mechanism can suppress the influence of conflicting gradient directions across tasks and maintain the consistency of model parameter updates, which is particularly suitable for multi-task fine-tuning or sequential task learning scenarios. By introducing the above-mentioned gradient perception mechanism, the method as a whole can achieve parameter adjustments with reasonable structure, clear direction, and moderate strength under extremely low samples, thereby improving fine-tuning efficiency and stability.

## III. EXPERIMENT

### A. Datasets

This study adopts the SuperGLUE dataset as the benchmark for fine-tuning and evaluating large language models. The goal is to assess the proposed method's adaptability under few-shot conditions. SuperGLUE is a comprehensive benchmark for natural language understanding. It includes a variety of challenging tasks such as textual entailment, question answering, and causal reasoning. The dataset is known for its linguistic diversity and task complexity.

Tasks in the dataset include, but are not limited to, BoolQ, CB, COPA, MultiRC, RTE, and WiC. Each task provides a predefined training set, validation set, and test set. These allow the evaluation of different language understanding abilities. In this study, only a small portion of the training samples is used for each task. This creates a few-shot learning setting to test the effectiveness of the fine-tuning strategy under data-scarce conditions.

The SuperGLUE data is standardized. The input format is unified as one or more pairs of text sequences. The output is either multi-class labels or binary decisions. The dataset is well-structured, with clear task boundaries and standard evaluation metrics. This ensures a fair comparison across methods and supports a realistic and controlled environment for few-shot fine-tuning experiments.

### B. Experimental Results

First, the comparative experimental results are given, and the experimental results are shown in Table 1.

Table 1. Comparative experimental results

| Method | Avg Accuracy(%) | Gradient Stability | Directional Alignment |
|---|---|---|---|
| Full Fine-tuning[19] | 74.3 | 0.61 | 0.52 |
| Adapter-Tuning[20] | 76.5 | 0.67 | 0.58 |
| LoRA[21] | 77.2 | 0.69 | 0.60 |
| Prompt-Tuning[22] | 78.0 | 0.71 | 0.61 |
| Ours | 80.1 | 0.78 | 0.73 |

The experimental results demonstrate that the proposed gradient-based fine-tuning method surpasses several prominent techniques in various crucial metrics. It achieves an impressive average accuracy of 80.1%, substantially surpassing conventional methods like full fine-tuning and adapter-tuning. This finding suggests that, under limited data constraints, relying solely on full-parameter or submodule tuning fails to fully harness the potential of large language models. Conversely, integrating gradient information offers more effective guidance for task-specific adaptation.

In terms of gradient stability, the proposed method also demonstrates a clear advantage. It achieves a stability score of 0.78, reflecting lower gradient fluctuations compared to conventional methods. This suggests that the gradient-driven optimization strategy offers a smoother update path in the parameter space. It helps prevent performance degradation caused by unstable update directions in few-shot settings. Compared with full fine-tuning, which scores 0.61, the proposed approach better maintains dynamic consistency during training.

For directional alignment, the method achieves a score of 0.73, which is noticeably higher than that of competing models. This improvement confirms the effectiveness of the directional regularization term. It aligns the model gradients with task priors or target directions during optimization. As a result, the method enhances both the precision and efficiency of fine-tuning. In contrast, approaches like LoRA and prompt-tuning, which lack this mechanism, perform more conservatively on this metric. In summary, the experimental results confirm that the gradient-based optimization strategy improves performance while also enhancing the controllability and robustness of the fine-tuning process. This makes the method especially suitable for data-scarce scenarios or applications sensitive to update paths. It significantly improves the practicality and reliability of large language models under low-resource conditions. Combined with the theoretical design and methodological structure described earlier, this study provides a more interpretable and actionable optimization framework for few-shot fine-tuning.

This paper also presents a test of the model's migration and generalization capabilities under different sample sizes, and the experimental results are shown in Figure 2.

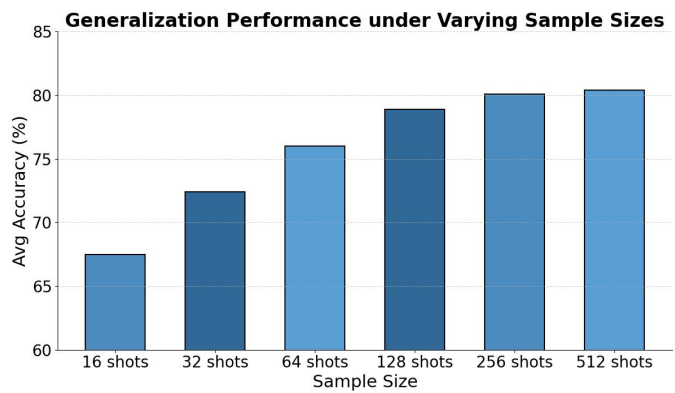

Figure 2. Test of the model's migration and generalization ability under different sample sizes

Figure 2 illustrates that average accuracy improves monotonically with increasing sample size, confirming the strong transfer and generalization capabilities of our gradient-based fine-tuning in few-shot scenarios. Even with as few as 16 or 32 examples, the model attains reasonable accuracy, demonstrating rapid initial adaptation. The most pronounced gains occur between 64 and 128 samples—where precise control over gradient direction and magnitude accelerates convergence toward task objectives—whereas beyond 256 samples, accuracy plateaus, indicating that the model has effectively internalized core distributional features. These trends validate that gradient-guided updates enable efficient convergence and robust performance in low-resource settings, making the method well suited for real-world applications with limited training data. This paper also presents an adaptability test of gradient optimization fine-tuning in specific domain tasks, and the experimental results are shown in Figure 3.

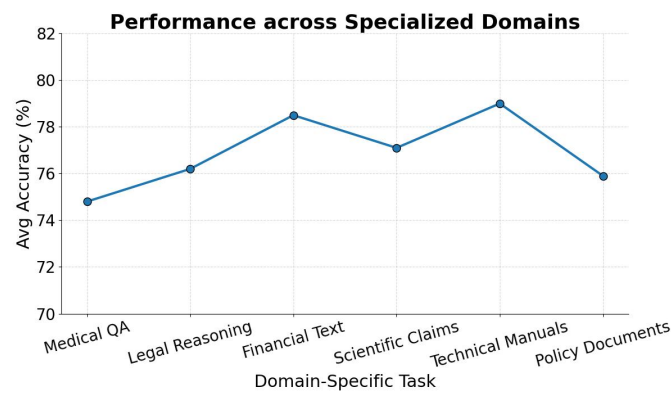

Figure 3. Adaptability testing of gradient optimization fine-tuning on domain-specific tasks

As shown in the Figure 3, the proposed gradient-based fine-tuning method demonstrates strong adaptability across various domain-specific tasks. The accuracy remains consistently above 75%. In particular, the method shows strong generalization on financial texts and technical manuals, reaching 78.5% and 79.0% accuracy respectively. This indicates that in structurally complex and terminology-rich texts, the gradient-based control of direction and magnitude helps the model capture core semantics, enabling stable and effective fine-tuning.

On tasks involving legal reasoning and scientific statements, the model also achieves performance close to 77%. This confirms the robustness of the method in logical inference and causal understanding. The stability is attributed to the modeling of gradient direction consistency. This helps align parameter updates with task objectives in the parameter space, reducing generalization error. Compared with traditional methods that rely on large datasets to learn task-specific features, this approach can locate effective feature spaces with fewer samples.

It's worth noting that the policy document task exhibits slight fluctuations in model performance, with accuracy dropping to 75.9%. This could be attributed to the high stylistic variability and semantic ambiguity inherent in this type of text. Even with gradient optimization, challenges persist in handling domains with unstable expression patterns. This underscores the necessity for future integration of domain adaptation mechanisms to enhance gradient constraints and improve adaptability in such intricate tasks. Overall, the experimental results demonstrate the robust transferability and adaptability of the proposed method across multiple specialized fields. It maintains stable optimization paths under diverse task structures. These findings further emphasize the pivotal role of gradient information in few-shot fine-tuning. Especially in real-world applications with varied contexts and complex task requirements, this method presents a practical approach for constructing resilient language models. Additionally, this paper presents a loss function drop graph, as depicted in Figure 4.

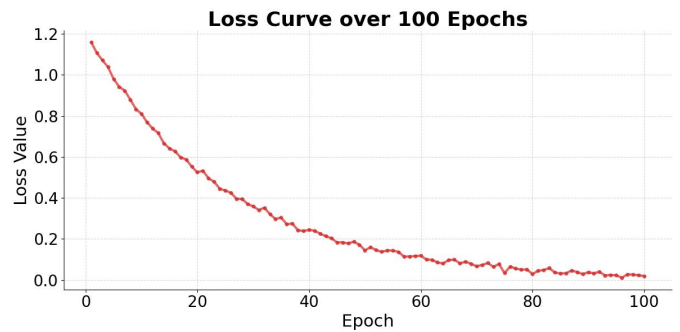

Figure 4. Loss function drop graph

From the loss curve, it's evident that the model experiences a rapid decline in loss during the initial training phase (the first 20 epochs). This suggests that the incorporation of gradient-based optimization facilitates the model's swift identification of effective parameter directions and its gradual alignment with the task. Notably, this rapid convergence is particularly crucial in few-shot settings, as it enables substantial optimization within a limited number of updates. This approach effectively mitigates the training bottleneck caused by data scarcity.

In the middle stage (approximately epochs 20 to 60), the loss continues to decline steadily, reflecting strong training stability. This period is when gradient regularization components, such as directional alignment and magnitude control, begin to take effect. By aligning parameter updates with task objectives, the model gradually improves its structural adaptation to the task. This shows the robustness of the gradient optimization method in handling complex learning scenarios without heavy data reliance.

In the later stage (after 60 epochs), the loss gradually converges and becomes more stable. This suggests that the model has effectively learned the distribution of the target task. The smooth convergence process indicates that the model is nearing an optimal region in parameter space. There are no signs of instability or severe overfitting. This indirectly confirms the effectiveness of gradient stability and fine-tuning path control, demonstrating the method's ability to maintain reliable convergence at the end of training. Overall, the loss curve provides strong evidence that the proposed gradient optimization method maintains clear task-oriented guidance and control throughout all training phases. Through stable loss compression, the model not only achieves strong performance under few-shot conditions but also retains efficient and robust training behavior. This offers a solid foundation for future deployment and generalization.

## IV. Conclusion

This paper proposes a gradient-informed fine-tuning method for large language models under few-shot conditions. By introducing mechanisms for gradient direction consistency and magnitude control, the method enables fine-grained regulation of parameter update paths. Under low-resource settings, it effectively improves task adaptability and training stability, demonstrating strong performance and generalization. Compared to traditional fine-tuning strategies, this approach avoids large-scale full-parameter updates and structural modifications, offering higher efficiency and broader applicability. Experimental results across multiple dimensions show that the method maintains stable optimization even in data-scarce and domain-shifting environments. It performs well on professional language tasks such as legal, medical, and financial applications. This capability reduces dependence on large datasets and provides a practical solution for building low-cost and highly adaptable intelligent systems. In real-world scenarios where rapid model transfer and deployment are required, the proposed approach offers significant advantages.

The gradient-aware mechanism introduced in this method also opens new perspectives for future research in multi-task learning, continual learning, and model safety. As models grow larger and application demands become more diverse, achieving fine control over parameter updates without sacrificing performance will be key to making language models more practical. Future work can explore the interaction between gradient signals and internal model structures to develop more interpretable and adaptive fine-tuning frameworks. Extending this method to multimodal tasks or cross-lingual settings and evaluating its generalization in complex semantic environments will be both challenging and promising. The ultimate goal is to enable large language models to perform more efficiently, controllably, and reliably across a wide range of real-world tasks.